\documentclass[runningheads]{llncs}
\usepackage[T1]{fontenc}
\usepackage{amsmath,amssymb}
\usepackage{graphicx}
\authorrunning{C. Petrou et al.}

\begin{document}
\title{Gaze Prediction in Virtual Reality Without Eye Tracking Using Visual and Head Motion Cues}
\author{
Christos Petrou\inst{1} \and
Harris Partaourides\inst{2} \and
Athanasios Balomenos\inst{1} \and
Yannis Kopsinis\inst{1} \and 
Sotirios Chatzis\inst{3}
}

\institute{
Libra AI Technologies, Athens, Greece\\
\and
Ethical AI Novelties, Limassol, Cyprus\\
\and
Cyprus University of Technology, Limassol, Cyprus
}
\maketitle

\begin{abstract}
Gaze prediction plays a critical role in Virtual Reality (VR) applications by reducing sensor-induced latency and enabling computationally demanding techniques such as foveated rendering, which rely on anticipating user attention. However, direct eye tracking is often unavailable due to hardware limitations or privacy concerns. To address this, we present a novel gaze prediction framework that combines Head-Mounted Display (HMD) motion signals with visual saliency cues derived from video frames. Our method employs UniSal, a lightweight saliency encoder, to extract visual features, which are then fused with HMD motion data and processed through a time-series prediction module. We evaluate two lightweight architectures, TSMixer and LSTM, for forecasting future gaze directions. Experiments on the EHTask dataset, along with deployment on commercial VR hardware, show that our approach consistently outperforms baselines such as Center-of-HMD and Mean Gaze. These results demonstrate the effectiveness of predictive gaze modeling in reducing perceptual lag and enhancing natural interaction in VR environments where direct eye tracking is constrained.

\end{abstract}

\section{Introduction}

Gaze estimation, the process of inferring a user’s visual attention, is a cornerstone of human–machine interaction~\cite{Duchowski2017,Hansen2010}. By providing a continuous, non-intrusive measure of attention, it enables adaptive interfaces that respond naturally to user intent in domains such as assistive technologies~\cite{griffiths2024use}, usability assessment~\cite{novak2023eye,wang2019exploring}, and interactive media~\cite{li2024real}. In Virtual Reality (VR), gaze estimation is particularly crucial: it supports natural interaction in immersive 3D environments and powers performance-optimizing techniques such as foveated rendering, where computational resources are concentrated on the user’s attended region~\cite{Ruhland2015}.

Going beyond estimation, gaze prediction anticipates upcoming attentional targets. This capability reduces perceptual lag by compensating for sensor and rendering delays~\cite{Stein2020_EyeTrackingLatency}, while also enabling pre-allocation of computational resources to likely regions of interest. As a result, prediction not only improves responsiveness in dynamic scenes but also facilitates more efficient and adaptive VR systems.

Despite its promise, gaze prediction in VR is often constrained by limited access to eye-tracking data. Hardware restrictions and privacy regulations frequently prevent developers from accessing raw gaze streams; for instance, the Apple Vision Pro (AVP) exposes only filtered, privacy-preserving gaze data. Such limitations introduce latency and processing bottlenecks, degrading responsiveness and reducing the effectiveness of attention-aware applications.

To overcome these challenges, we propose a multimodal gaze prediction framework that operates without direct eye-tracking input. Our method combines Head-Mounted Display (HMD) motion signals with visual saliency cues derived from recent frames. A lightweight UniSal-based saliency encoder~\cite{Droste2020} extracts content-driven attention features, which are fused with HMD motion data and passed to a sequence encoder with a time-series prediction module. We evaluate two architectures, TSMixer~\cite{Chen2023} and LSTM~\cite{Hochreiter1997}, for their ability to forecast both short-term and long-term gaze shifts.

\begin{figure}[t]
\centering
\includegraphics[width=1\textwidth]{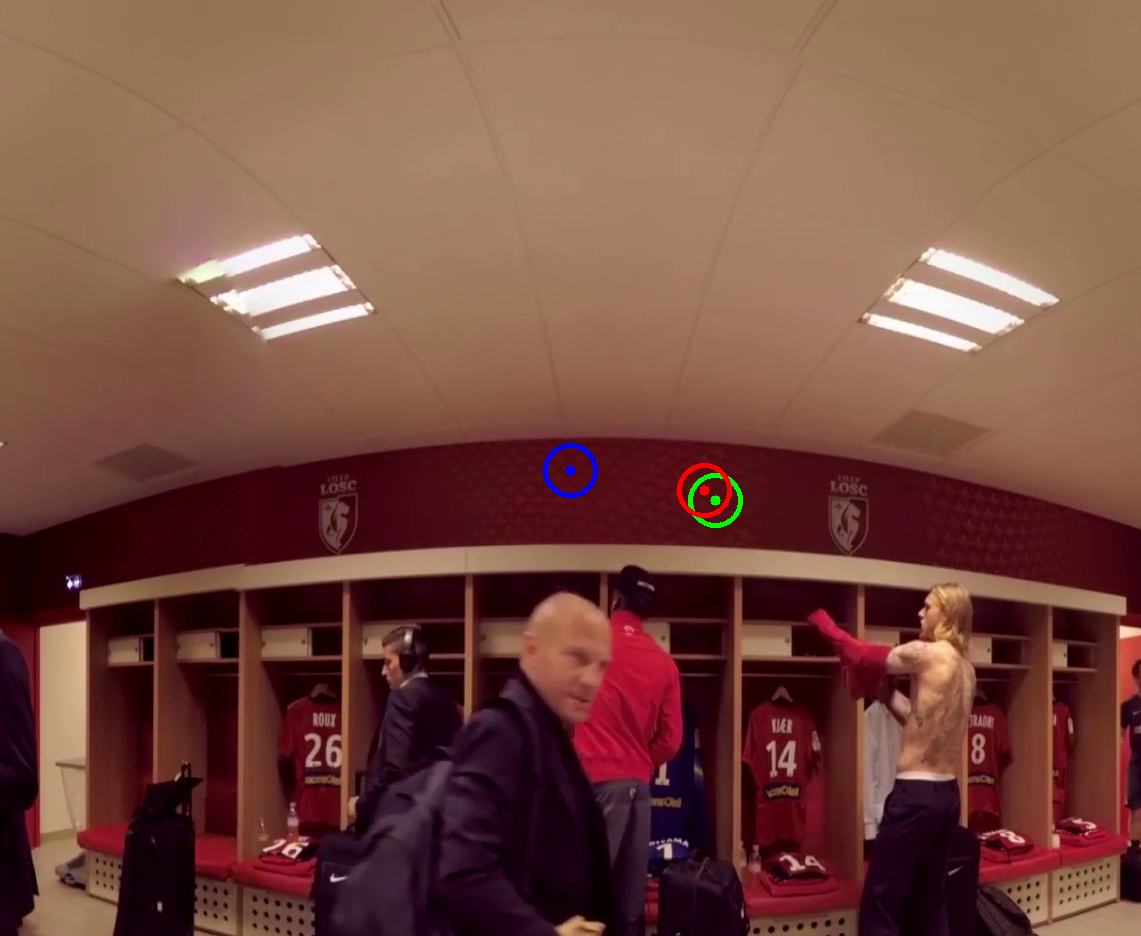}
\caption{Example of gaze prediction on the EHTask dataset. Blue: HMD center; green: true gaze; red: predicted gaze using the proposed method.}

\label{fig:EHT_sample}
\end{figure}

By predicting gaze trajectories in advance, our approach mitigates the impact of restricted or delayed eye-tracking streams, reducing latency and improving responsiveness on privacy-sensitive and resource-limited hardware. Figure~\ref{fig:EHT_sample} illustrates an example prediction from the EHTask~\cite{Hu2023} dataset, comparing HMD center, true gaze, and predicted gaze. We validate our method on the EHTask benchmark and implement it across both Linux-based environment and the Apple ecosystem. Results show that our framework consistently outperforms baseline strategies such as Center-of-HMD and Mean Gaze, while capturing complex spatiotemporal gaze dynamics in immersive VR scenarios.

The main contributions of this work are:

\begin{itemize}
\item A novel multimodal gaze prediction framework that integrates visual saliency with real-time HMD motion, eliminating reliance on dedicated eye-tracking hardware.
\item A comprehensive evaluation on the EHTask benchmark, demonstrating accurate short-term gaze prediction at 333 ms and extending the prediction horizon up to 1s, thereby characterizing the temporal limits of reliable forecasting.
\item A cross-platform implementation on Linux-based systems and within the Apple ecosystem, highlighting practical applicability under privacy and hardware constraints while supporting real-time gaze trajectory prediction.
\end{itemize}

Through these contributions, we advance gaze prediction as a practical and deployable solution for enhancing VR performance and interaction in next-generation head-mounted systems.

\section{Related Work}

Estimating or predicting gaze in VR without direct eye-tracking access typically relies on indirect cues or predictive models. A common baseline is head-orientation approximation~\cite{Hu2022,Pfeiffer2012}, which assumes that users look in the direction their head is facing. While simple and computationally efficient, this approach fails to capture independent eye movements. To improve upon this, head–gaze correlation models~\cite{Tong2017} statistically learn the relationship between head orientation and eye dynamics, offering greater accuracy by modeling their joint behavior.

Beyond head-based signals, content-aware methods exploit visual scene information to anticipate gaze. Task- and context-driven models~\cite{Li2023} incorporate user goals and actions, while saliency-based approaches~\cite{Borji2013,Hu2020,Jiang2015SALICON,Kummerer2017} predict attention by identifying visually prominent regions such as areas of high contrast, motion, or semantic importance. Among these, UniSal stands out as a lightweight yet effective saliency predictor. Built on a MobileNetV2 backbone~\cite{Sandler2018}, UniSal leverages separable depthwise convolutions to minimize parameters and floating-point operations while retaining rich feature representations. Its multi-scale fusion design captures both fine-grained details and global contextual cues, and its hybrid training objective, combining Kullback–Leibler divergence, Pearson’s correlation coefficient, and normalized scanpath saliency, ensures strong alignment with diverse evaluation metrics. These properties enable UniSal to generalize across datasets and achieve near real-time performance, addressing the limitations of traditional saliency models that neglect temporal gaze dynamics or require explicit eye-tracking signals, which may pose privacy concerns.

Motivated by these constraints in emerging VR platforms such as the Apple Vision Pro, we propose a multimodal approach that fuses HMD motion with UniSal-derived saliency cues to estimate and forecast gaze without direct eye access. Temporal dynamics are captured through lightweight predictors, either TSMixer or LSTM, each independently evaluated for their ability to model gaze trajectories over time. Compared to head-forward baselines or correlation-only methods, our approach achieves more accurate and anticipatory gaze prediciton, making it particularly well-suited for privacy-sensitive, resource-constrained VR applications.

\section{Methodology}

The objective of this work is to establish a gaze prediction framework that eliminates the need for dedicated eye-tracking hardware in head-mounted devices such as the AVP. Our approach combines HMD positional signals with visual saliency cues to enable accurate and robust gaze estimation. The framework consists of two core modules, the Saliency Encoder and the Sequence Encoder. The Saliency Encoder extracts attention-relevant visual features, while the Sequence Encoder incorporates these features along with real-time HMD motion information to forecast future gaze directions.

The Saliency Encoder processes visual frames to extract saliency maps, which highlight attention-relevant areas. For this, we utilize the lightweight UniSal network. The choice of UniSal aligns with the requirements of quick generation of saliency maps, which is critical for near real-time VR applications.

\begin{figure}[t]
\centering
\includegraphics[width=1\textwidth]{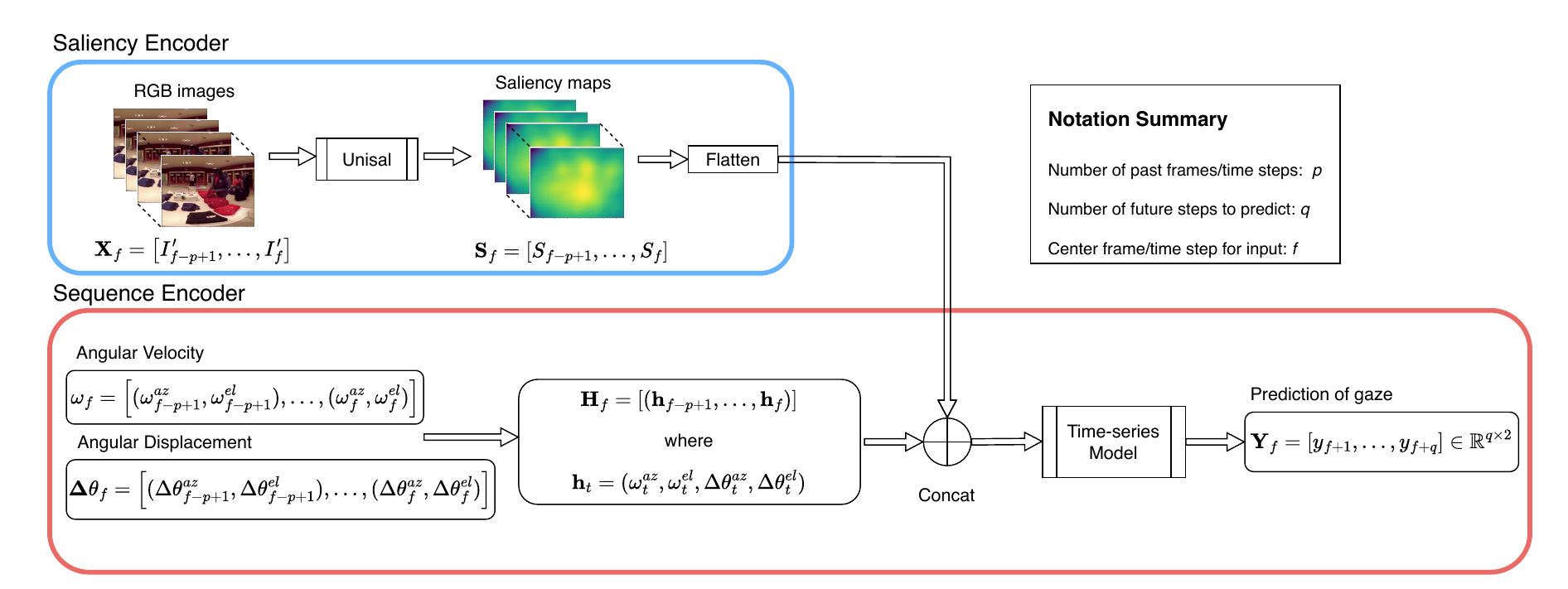}
\caption{Overview of the multi-modal gaze prediction process}
\label{fig:architecture}
\end{figure}

The Sequence Encoder models temporal patterns in head motion and visual saliency over a fixed-length window of recent frames. At each time step, the 3D positional vector of the HMD is recorded and pre-processed on the fly to compute motion-related features, such as linear velocity and angular displacement, which are closely correlated with gaze shifts. The saliency maps produced by the Saliency Encoder are flattened into feature vectors and fused with these HMD motion features. The combined multi-modal sequence is then passed through the time-series prediction module within the Sequence Encoder, which predicts future gaze directions based on the recent patterns of visual attention and head motion. An overview of the multi-modal gaze prediction process is presented at Figure \ref{fig:architecture}.

To train the model, we utilize a dataset of synchronized video sequences and HMD motion metadata, allowing the system to learn the relationships between visual saliency, head motion, and gaze trajectories. Each video $V$ is paired with a corresponding metadata file containing frame-wise annotations of head orientation and gaze direction. At each valid time step, input–output pairs are generated by combining past observations with future gaze targets. This procedure is applied consistently during both training and inference, ensuring the model effectively captures temporal dependencies in the data.

\subsection{Multi-Modal Data Preparation}

To define valid sampling points, let $T$ denote the total number of frames in a video. For a given frame index $f$ such that
\begin{equation}
f \in \{ f \mid p \le f \le T - q \},
\end{equation}
where $p$ is the number of past frames and $q$ is the number of future prediction steps, we define the input-output sample centered at frame and time point $f$. This ensures that each sample has sufficient past context and a well-defined future horizon.

For each sample, we construct input features by combining visual information and head motion cues. For the visual modality, we extract a sequence of $p$ preprocessed RGB frames
\begin{equation}
\mathbf{X}_f = \left[ I_{f - p + 1}', \dots, I_{f}' \right],
\end{equation}
where $I_t' = g(I_t)$ is obtained via a preprocessing function $g(\cdot)$, such as resizing and normalization. The resulting tensor has shape $\mathbf{X}_f \in \mathbb{R}^{p \times H \times W \times 3}$. Similarly, for each corresponding frame, we compute the saliency heatmap
\begin{equation}
\mathbf{S}_f = \left[ S_{f - p + 1}, \dots, S_f \right],
\end{equation}
where $S_t = s(I_t')$ is the output of the saliency module applied to the preprocessed frame $I_t'$. The saliency tensor has shape
\begin{equation}
\mathbf{S}_f \in \mathbb{R}^{p \times H \times W \times 1}.
\end{equation}
This single-channel heatmap highlights regions of visual importance for each frame and can be used for gaze prediction.

In addition to visual frames, head motion features are computed for each input frame $t$ as
\begin{equation}
\mathbf{h}_t = \left( \omega_t^{az}, \omega_t^{el}, \Delta\theta_t^{az}, \Delta\theta_t^{el} \right),
\end{equation}
where $\omega_t$ denotes angular velocity and $\Delta\theta_t$ represents angular displacement in azimuth and elevation, derived from the headset orientation $\boldsymbol{\theta}^{\text{HMD}}_t$ using a spherical motion model. The full sequence of motion features for the input window is then
\begin{equation}
\mathbf{H}_f = \left[ \mathbf{h}_{f - p + 1}, \dots, \mathbf{h}_f \right] \in \mathbb{R}^{p \times 4}.
\end{equation}
The saliency features and head motion features are concatenated to form the joint input representation
\begin{equation}
\mathbf{T}_f = \text{concat}\left( \mathbf{S}_f, \mathbf{H}_f \right).
\end{equation}
This joint representation is then processed the time-series model $\mathcal{T}(\cdot)$, which captures temporal dependencies across the input window and produces the predicted gaze sequence:
\begin{equation}
\mathbf{Y}_f = \mathcal{T}(\mathbf{T}_f).
\end{equation}
The target output for each sample is the gaze direction relative to the headset orientation. At time $f + t$, the ground-truth (GT) gaze is
\begin{equation}
\mathbf{y}_{f + t} = \left( \delta\theta^{az}_{f + t}, \delta\theta^{el}_{f + t} \right),
\end{equation}
with angular offsets computed as
\begin{equation}
\delta\theta^{az}_{f + t} = \text{wrap}\left( \theta^{\text{GT}}_{az}(f + t) - \theta^{\text{HMD}}_{az}(f) \right),
\end{equation}
\begin{equation}
\delta\theta^{el}_{f + t} = \theta^{\text{GT}}_{el}(f + t) - \theta^{\text{HMD}}_{el}(f),
\end{equation}
where the wrap function normalizes angles to $[-180^\circ, 180^\circ]$:
\begin{equation}
\text{wrap}(\delta) = 
\begin{cases}
\delta - 360^\circ & \text{if } \delta > 180^\circ, \\
\delta + 360^\circ & \text{if } \delta < -180^\circ, \\
\delta & \text{otherwise}.
\end{cases}
\end{equation}
The full prediction sequence is then
\begin{equation}
\mathbf{Y}_f = \left[ \mathbf{y}_{f + 1}, \dots, \mathbf{y}_{f + q} \right] \in \mathbb{R}^{q \times 2}.
\end{equation}
Each training sample is represented as a triplet $(\mathbf{X}_f, \mathbf{H}_f, \mathbf{Y}_f)$, and $N$ such samples are aggregated across one or more videos into minibatches of size $m$:
\begin{equation}
\mathcal{B} = \left\{ \left( \mathbf{X}_1, \mathbf{H}_1, \mathbf{Y}_1 \right), \dots, \left( \mathbf{X}_m, \mathbf{H}_m, \mathbf{Y}_m \right) \right\}.
\end{equation}
Throughout this formulation, $p$ denotes the number of past frames, $q$ the number of future steps to predict, $H \times W$ the spatial resolution of input frames, $m$ the minibatch size, and $\theta^{az}, \theta^{el}$ represent azimuth and elevation angles in degrees.

\subsection{Time-Series Prediction Model Training}

In order to train the time-series prediction model on the constructed dataset, we employ a loss function specifically designed to handle angular outputs, ensuring accurate modeling of both azimuth and elevation gaze components. To this end, we define a Spherical MSE Loss function tailored to azimuth and elevation outputs. Let $\hat{\mathbf{y}} = (\hat{\theta}_{az}, \hat{\theta}_{el})$ denote the predicted angles and $\mathbf{y} = (\theta_{az}, \theta_{el})$ the ground truth, all expressed in degrees. The azimuth difference $\Delta_{az}$ is computed as the absolute difference between predicted and true azimuths, taking into account angle normalization, while the elevation difference $\Delta_{el}$ is computed as the linear difference between predicted and true elevations, reflecting the bounded range of this angle:
\begin{equation}
\Delta_{az} = |\hat{\theta}_{az} - \theta_{az}|, \quad
\Delta_{el} = |\hat{\theta}_{el} - \theta_{el}|.
\end{equation}
From these, we define the mean angular loss
\begin{equation}
\mathcal{L}_{\text{angular}} = \frac{1}{2} (\Delta_{az} + \Delta_{el}),
\end{equation}
and the Spherical MSE Loss
\begin{equation}
\mathcal{L}_{\text{MSE}} = \frac{1}{2} (\Delta_{az}^2 + \Delta_{el}^2),
\end{equation}

\section{Experiments}

We evaluate our gaze prediction approach using the EHTask dataset, a publicly available collection of 15 curated 360-degree videos with precise eye and head movement recordings from 30 participants. Each participant experienced four experimental conditions: \textit{Free Viewing} (watching without any specific goal, serving as a baseline), \textit{Visual Search} (actively looking for a target object in the scene), \textit{Saliency} (focusing on the most visually prominent elements), and \textit{Track} (continuously following a moving object).

To prepare the data for training and evaluation, each 2-minute recording is divided into non-overlapping 5-second segments sampled at 30 Hz. From each segment, we extract 15 frames to provide past context and 10 frames for future gaze prediction. A sliding window with a step size of 5 frames is applied, generating approximately 8 training samples per segment and yielding 69,120 samples across the dataset. An example of gaze prediction from the EHTask dataset is shown at Figure \ref{fig:EHT_sample}.

A custom data loader constructs three main tensors for each batch:

\begin{itemize}
    \item Saliency Input: $[\text{batch} \times p \times 3 \times 288 \times 384]$, representing RGB frames.
    \item Motion Sequence: $[\text{batch} \times p \times 4]$, containing angular velocity and displacement features.
    \item Target Sequence: $[\text{batch} \times q \times 2]$, representing future gaze offsets relative to the headset orientation.
\end{itemize}

For each frame, saliency maps are generated using a pre-trained UniSal model, normalized to the range $[0,1]$, passed through convolutional layers, and subsequently flattened before being integrated with motion features to construct the input sequence for the temporal gaze prediction model.

For training and evaluation, the dataset is divided into an 80/20 split for training and validation, with additional held-out videos reserved for testing. Optimization is performed using the Adam optimizer with a learning rate of 0.001 and a batch size of 32. Early stopping is applied, terminating training after 8 consecutive epochs without improvement on the validation set.

Evaluation Metrics: Model performance was assessed using the Spherical Mean Squared Error (SphericalMSELoss). The evaluation was performed on a test set of six designated videos. Performance metrics included both Spherical MSE and Root Mean Squared Error (RMSE), computed for each axis and for each future prediction timestep. To provide context, results were compared against a center-of-HMD baseline, which assumes gaze aligns with the HMD's forward direction.

\subsection{Temporal Sequence Modeling}
To model temporal dependencies in gaze dynamics, we evaluated two distinct sequence modeling architectures: the TSMixer, a convolutional temporal mixer designed for efficient spatiotemporal feature learning, and a Long Short-Term Memory (LSTM) network, a recurrent architecture capable of capturing long-range temporal correlations. This comparison allows us to assess the effectiveness of both modern convolutional mixers and traditional recurrent models for predicting future gaze trajectories from past visual and motion information.

To evaluate the predictive capabilities of the developed models, experiments were conducted for both short-term and long-term gaze forecasting. Specifically, predictions were made for 10 timesteps into the future (corresponding to approximately x`3 ms), as well as for 30 timesteps (corresponding to 1000 ms).

\begin{table}
\centering
\caption{Comparison of models based on Spherical RMSE.}
\begin{tabular}{|c|c|}
\hline
\textbf{Model} & \textbf{Spherical RMSE} \\
\hline
LSTM & \textbf{5.166} \\
\hline
TSMixer & 5.314 \\
\hline
Center HMD & 10.106 \\
\hline
Mean HMD & 10.548 \\
\hline
\end{tabular}
\label{tab:rmse_comparison}
\end{table}

\begin{figure}[t]
\centering
\includegraphics[width=0.95\textwidth]{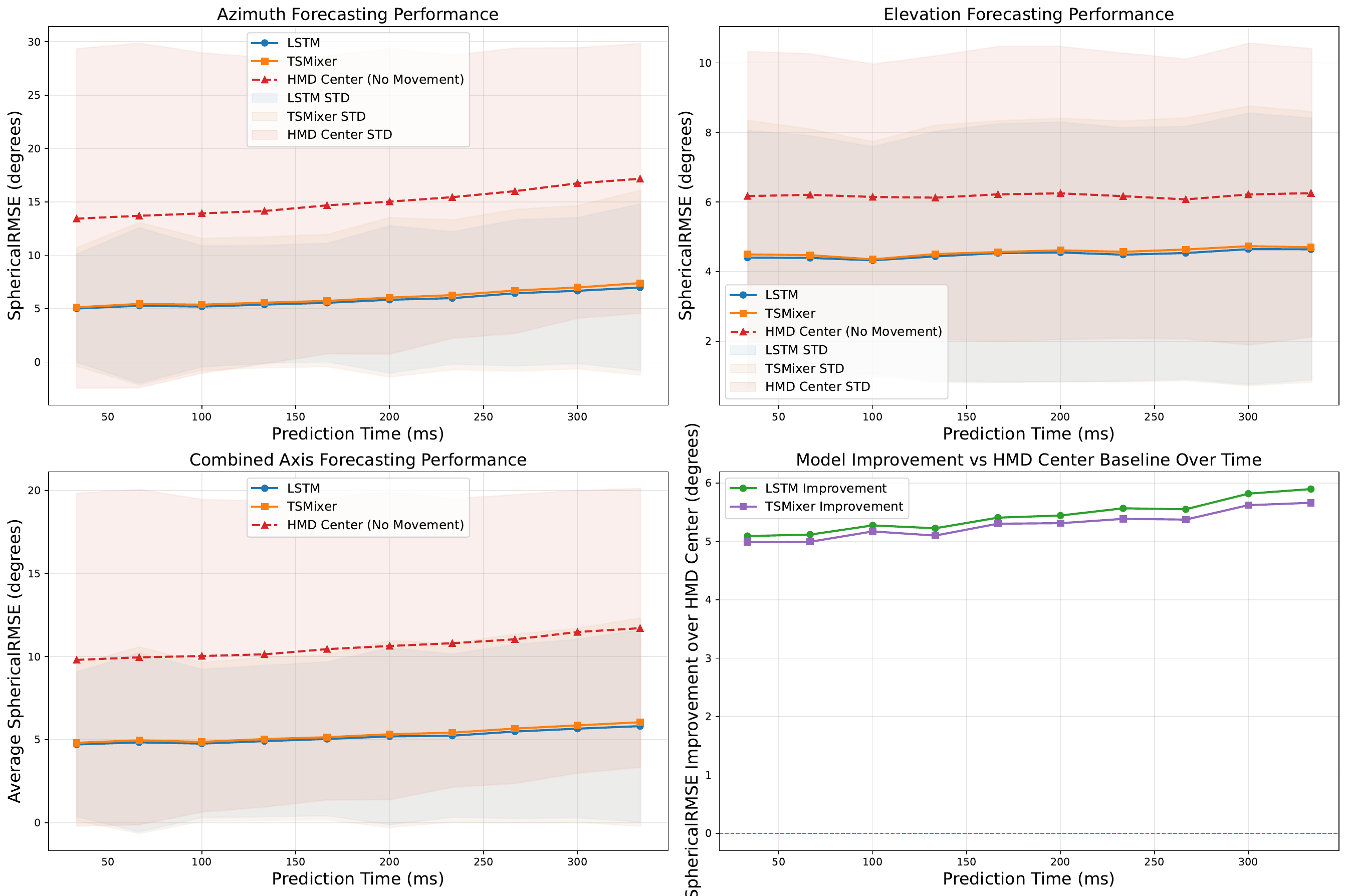}
\caption{Test set evaluation of gaze prediction over a 333 ms horizon. Results for LSTM and TSMixer are compared with the center-of-HMD baseline: azimuth, elevation, combined axes, and relative improvement over the baseline.}
\label{model_eval_10}
\end{figure}

Figure~\ref{model_eval_10} presents the Spherical RMSE and standard deviation errors on the test set for the 10-timestep prediction horizon. The figure consists of four subplots, reporting errors and standard deviations along the horizontal axis, the vertical axis, their combined measure, and the relative improvement over the HMD center baseline. Both the LSTM and TSMixer implementations demonstrate substantially lower errors compared to the HMD center, while also accounting for variability, with performance differences reaching nearly 6 degrees at the final prediction steps. Table~\ref{tab:rmse_comparison} summarizes the Spherical RMSE values across all prediction horizons, confirming that both models consistently outperform the HMD center by approximately 5 degrees. For reference, the mean HMD baseline, computed as the average of past HMD center values, is also reported.

\begin{figure}[t]
\centering
\includegraphics[width=0.95\textwidth]{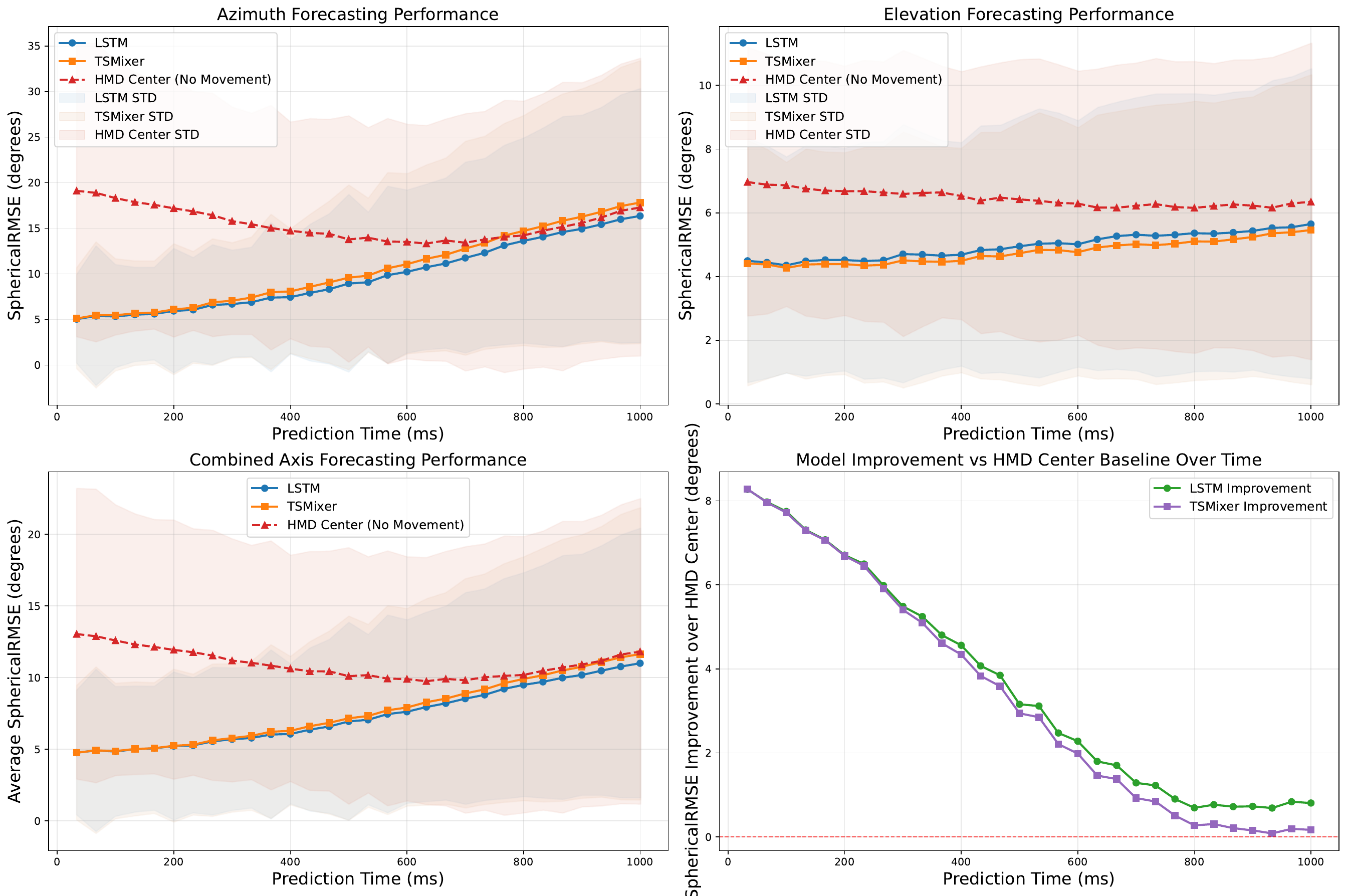}
\caption{Test set evaluation of gaze prediction over a 1000 ms horizon. Results for LSTM and TSMixer are compared with the center-of-HMD baseline: azimuth, elevation, combined axes, and relative improvement over the baseline.}
\label{model_eval_30}
\end{figure}

Figure~\ref{model_eval_30} shows the corresponding results for the 30-timestep prediction horizon. This experiment assesses the models’ ability to forecast gaze one second into the future, a challenging setting given the temporal distance. The results indicate that, in general, both models outperform the HMD center baseline. However, beyond 600 ms the errors converge toward those of the baseline, reducing the performance gap. Notably, after approximately 400 ms a clear difference emerges between the two models, with the LSTM achieving superior accuracy compared to TSMixer. Around this point, a distinct inflection (‘knee’) in the error curve is observed, suggesting that predictions beyond this horizon lose practical utility. These findings validate the choice of a 10-timestep prediction horizon as an effective and reliable setting for the proposed framework.

\subsection{Performance on Apple Ecosystem}

For deployment on the Apple Vision Pro (AVP), the trained model was converted into the CoreML format. The resulting CoreML model, combined with custom Swift-based pre- and post-processing components, constitutes the on-device gaze prediction pipeline. The final application is integrated into Unity, where HMD positional data are captured, processed by the CoreML model, and translated into predicted gaze directions for virtual reality enhancements.

Table~\ref{tab:processing_times} reports the execution times for different components of the pipeline, measured both on the Linux workstation used for model training (11th Gen Intel i9-11900KF @ 5.10 GHz, NVIDIA GeForce RTX 3090, 128 GB RAM) and on a MacBook Air with Apple Silicon (M1). The reported timings include per-image preprocessing, saliency model inference, and time-series prediction using the LSTM module. On the Linux workstation, the system achieves real-time performance. On the M1-based MacBook Air, real-time operation is feasible if saliency is not computed for every sequential frame; however, given the considerably higher performance of the M2 Apple Silicon in the AVP, the complete pipeline is expected to run in real time on the target device.

\begin{table}
\centering
\caption{Processing times for Linux and Apple Silicon.}
\begin{tabular}{|p{3cm}|p{3cm}|p{3cm}|p{2.5cm}|}
\hline
\textbf{Processing Time} & \textbf{Saliency Preproc.} \newline \textbf{per Image (ms)} & \textbf{Saliency Model} \newline \textbf{per Image (ms)} & \textbf{LSTM Model (ms)} \\
\hline
Linux Machine & 7.08 & 9.72 & 2.62 \\
\hline
Apple Silicon (M1) & 31.76 & 28.53 & 13.84 \\
\hline
\end{tabular}
\label{tab:processing_times}
\end{table}

\section{Conclusion}

This paper introduced a novel multi-modal gaze prediction framework for VR environments where direct eye-tracking is often unavailable due to hardware restrictions or privacy concerns. By eliminating the need for dedicated eye-tracking, our method achieves accurate gaze forecasting by effectively fusing real-time HMD motion signals with visual saliency cues extracted using a lightweight UniSal encoder. Evaluation on the EHTask dataset demonstrated that both the LSTM and TSMixer sequence modeling architectures consistently and substantially outperform baseline strategies such as Center-of-HMD and Mean Gaze. Specifically, the models provided accurate short-term gaze prediction within the reliable 333 ms prediction horizon, showing performance gains of nearly 5 degrees in Spherical RMSE over the HMD center baseline.

The successful cross-platform implementation on Linux and within the Apple ecosystem (using CoreML) further validates the framework's practical applicability and deployability on resource- and privacy-constrained head-mounted systems, such as the AVP. By anticipating user attention, our gaze prediction framework serves as a robust, indirect solution to mitigate perceptual lag, enable computationally demanding techniques like foveated rendering, and facilitate more natural interaction in immersive VR environments. These results confirm the utility of our multi-modal approach as a practical, deployable solution for enhancing VR performance in next-generation hardware.

\section{Acknowledgements}
This work was supported by the European Union’s Horizon Europe research and innovation programme under the Marie Skłodowska-Curie Actions (MSCA), Grant Agreement No. 101129865 (ACCESS).
%
%
\bibliographystyle{splncs04}
\bibliography{mybibliography}
\end{document}